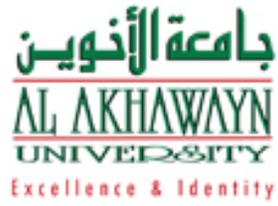

**Al Akhawayn University in Ifrane**

**School of Science and Engineering**

**The Lego Mindstorms Robotics Invention Systems 2.0 Toolkit:**

**A Study Case**

*by Ali Elouafiq*

*Supervised by Dr. Tajjedine Rachidi and Mr. Rachid El Ghoul*

## ABSTRACT:

This paper reviews the aspects of the LEGO® Mindstorms™ robotics invention system 2.0 ™ (RIS), by presenting the different elements of the kit, and relating them to actual robot components and norms. Furthermore a comparison between the LCS and Java is made, as well as comparing the RCX board to other technologies, specifically LEGO ® NXT and MIT's "Handy Board". Also, concrete examples of application using the RIS are presented.

## INTRODUCTION:

In order to make the initiation in robotics development easier, LEGO® developed the Mindstorms™ toolkit, with basic robotics concepts and principles that enable a person to develop easily a simple robot. This kit is widely used in most engineering institutions, such as MIT, since it enables and easy implementation of models. In this paper I'll present the different components of the robotics invention system kit, each component and its role in building a robot, and relate each one to actual robotic notations used worldwide, in order to make professional use of the kit; including the sensors, actuators and building blocks. In the Second section, the LEGO® script code, a C like language, will be briefly presented and will have some examples that will make its learning easier. Also, The LeJOS tiny Virtual Machine (a JVM) that could be installed on the RCX 2.0, that enables java programming on the RCX, will be presented with example. In the Last section, concrete application using the RIS will be presented; without forgetting that the RCX 2.0™ is a 10 years old technology, other concurrent technologies that are present in the current time, are more powerful and more flexible than the RCX 2.0™, since they enable multiple sensors and very fast computing and programming technology; notably, the LEGO® NXT 2.0™ and the MIT's "Handy Board", that enables multiple sensors and fast computation, besides the their large use and documentation .

## SECTION I: THE KIT

### I-A The Lego® Mindstorms™ Robotics Invention System ™2.0:

LEGO® Mindstorms is a line of programmable robotics/construction toys, manufactured by the LEGO® Group. Mindstorms originated from the programmable sensor blocks used in the line of educational toys. The first retail version of Lego Mindstorms was released in 1998 and marketed commercially as the Robotics Invention System (RIS). The hardware and software roots of the Mindstorms Robotics Invention System kit go back to the programmable brick created at the MIT Media Lab. This brick was programmed in Brick LOGO. [2]

The original Mindstorms Robotics Invention System kit contained two motors, two touch sensors, and one light sensor. Many kinds of real-life embedded systems, from elevator controllers to industrial robots, may be modeled using Mindstorms. [2]

Mindstorms kits are also sold and used as an educational tool, originally through a partnership between LEGO® and the MIT Media Lab .In addition, the shipped software can be replaced with third party firmware and/or programming languages, including some of the most popular ones used by professionals in the embedded systems industry, like *Java* and *C*. [2]

### I-B THE RCX 2.0:

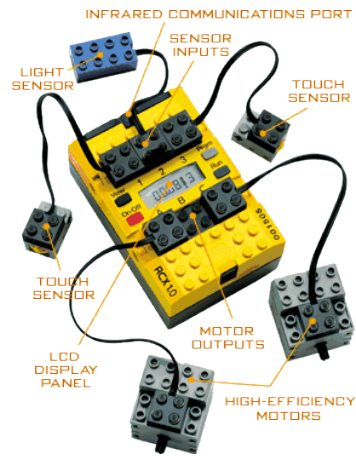

Figure I-B-1: RCX 2.0 Linked to different sensors and actuators.

(Figure credits to [1])

The first generation of Lego Mindstorms was built around a brick known as the RCX. It contains a 8-bit **Renesas H8/300 microcontroller** as its internal CPU. It also contains **32K of RAM** that stores the firmware and user programs. The brick is programmed by uploading a program from a Windows or Mac computer to the brick's RAM via a special infrared (IR) interface. After the user starts a program, an RCX-enabled Mindstorms creation may function totally on its own, acting on internal and external stimuli according to the programmed instructions. Also, two or several more RCX bricks can communicate with each other through the IR interface, enabling inter-brick cooperation or competition. In addition to the IR port, there are three sensor input ports and three motor output ports (also usable for other outputs). An LCD screen is available to battery level, the status of the input/output ports, which program is selected or running, and other information. [2]

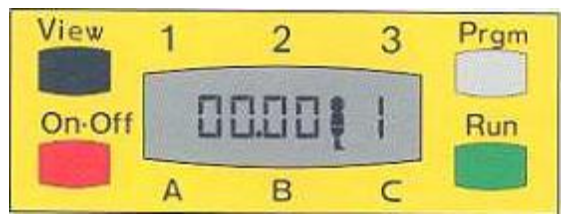

Figure I-A-1: RCX 2.0 Linked LCD Screen that displays the information

(Figure credits to [1])

In version 2.0 the power adapter jack was removed. The IR interface on the RCX is able to communicate with Spybots, Scout Bricks, Lego Train, and the NXT (using a third-party infrared link sensor.) .The RCX 2.0 IR carrier frequency is **76 kHz**. Both versions can transmit on either frequency. The carrier signal is generated by one of the RCX's internal timers. The RCX communicates with a computer using a Serial or USB IR tower. A patch is available for hyper-threading/multi-core CPUs. [2]

## I-C THE DIFFERENT BLOCKS AND PARTS OF THE MINDSTORMS TOOL KIT:

### I-C-1 LEGO INFRASTRUCTURE, MEASUREMENTS AND BRACING:

Plates are for structural reinforcement and spacing. While flat plates, smooth surfaces are for sliding mechanisms or for sensor and motor mounting

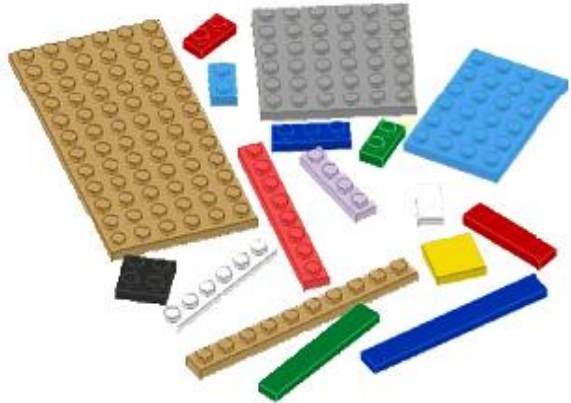

FIGURE I-C-1-A: LEGO dimensions (FLU).

(Figure credits to [4])

A Fundamental LEGO Unit (FLU) is equivalent to the length of a LEGO nub, as shown bellow:

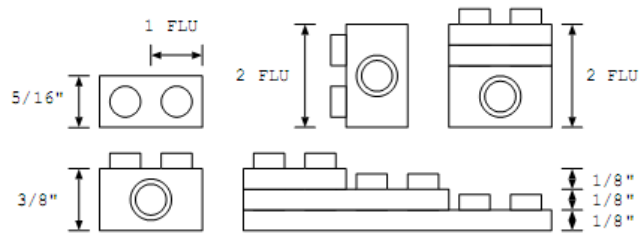

FIGURE I-C-1-B: LEGO dimensions (FLU).

(Figure credits to [3])

**Vertical Bracing:** To brace a Technic beam perpendicularly you need to sandwich two flat plates between two Technical beams, as shown in FIGURE I-C-1-C. Remember the beam-flat-flat-beam or "3-2-3 rule" makes bracing easier. There are several different vertical bracing arrangements possible. FIGURE I-C-1-D shows some unique examples. Although these examples work, they waste extra beams, and sometimes it can get hard keeping track of spacing. We highly recommended using the scheme shown in FIGURE I-C-1-C instead.[3]

**Diagonal Bracing:** Triangular trusses will only work if the number of nubs (FLUs) in the beam conforms to Pythagoras' theorem. Using Pythagoras' theorem not only leads to stronger bracing, but also saves beams. For the purposes of LEGO, this means that the only useful combinations are 3-4-5, 5-12-13, and 6-8-10 (Figure I-C-1-e).[3]

**Other Bracing:** Try experimenting with unconventional bracing. Many successful robots were based on designs that did not always have a perpendicular shape. Just remember that if there are angles in your chassis that are meant to be 90◦, not following basic trigonometry adds unnecessary shear forces to your chassis. This is especially important when it comes to the gearbox.[3]

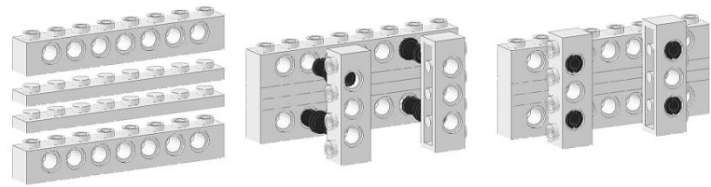

FIGURE I-C-1-C: Bracing: 1 beam = 3 flat plates

(Figure credits to [3])

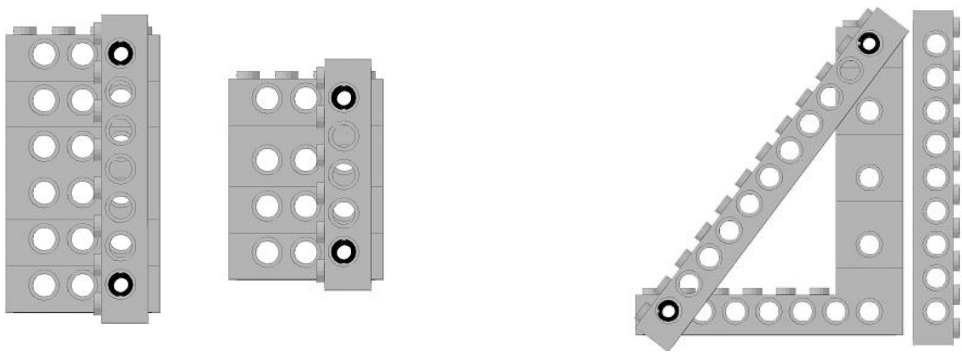

FIGURE I-C-1-D: Perpendicular bracing    Figure I-C-1-e: Pythagoras' Theorem for Diagonal Bracing

(Figures credits to [3])

### I-C-2 THE CONNECTORS:

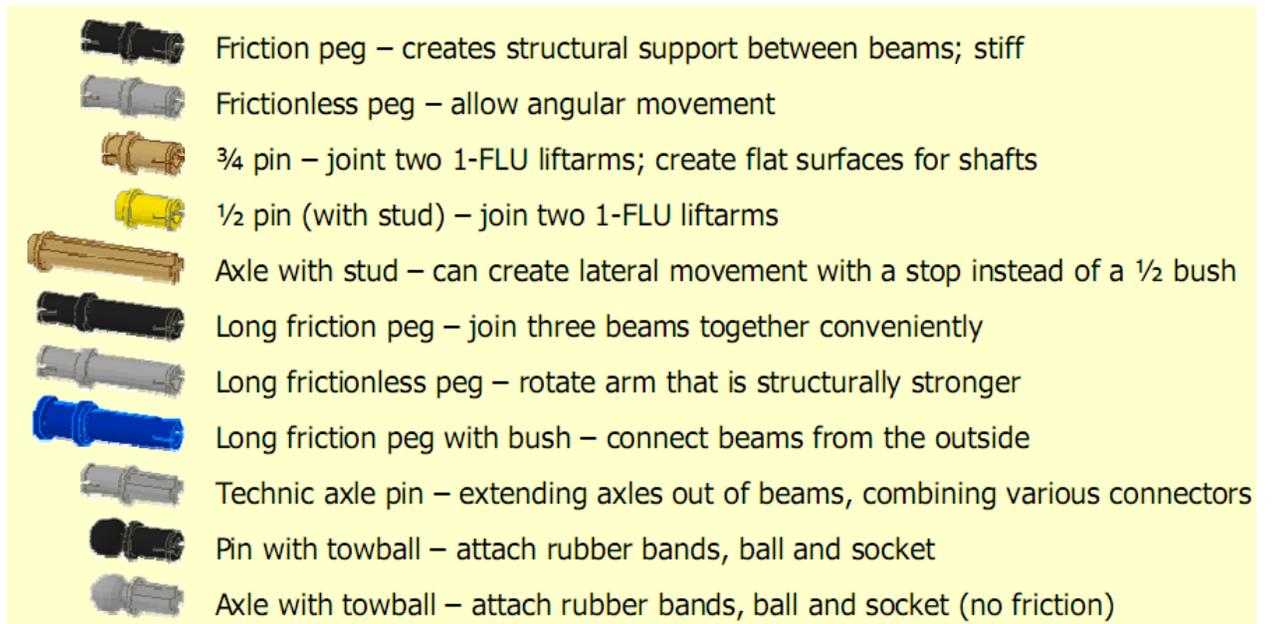

FIGURE I-C-2-A: The Different Connectors in the Lego Kit.

(Figures credits to [4])

### I-C-3 COMPARISON OF THE DIFFERENT ELEMENTS OF THE KIT

(Figures credits to [5])

To the untrained eye, Legos can be considered a very limited development environment: All the holes are predrilled, the lengths of the beams and axels predefined. But there are many other pieces in the set of Legos that give it versatility. [5]

**Comparison 1:**

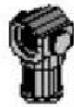

Connecting Rod        Angle Element, 0 degrees        Bushing/Catch Combo
2 FLUs tall           1 FLU + 3/8 inch tall           2 FLUs tall
*The axel inserted into this longer catch will grip better.*

**Comparison 2:**

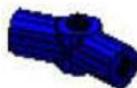        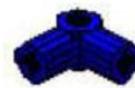

Angle Element, 157.5 degrees        Angle Element, 108 degrees
*Can create a 16 sided polygon*     *Can create a 5 sided polygon*
                                    *Can approximate curved surfaces*

**Comparison 3:** Lever Arms VS. Beams:

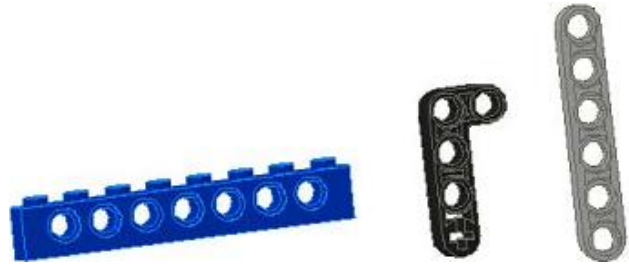

Lever Arms: Curved Ends, No Nubs, 2 different thicknesses: 1 FLU and ½ FLU

Uses: Brace gearboxes to allow for clearance, or to create odd angles

**Comparison 4:** Connector Pegs

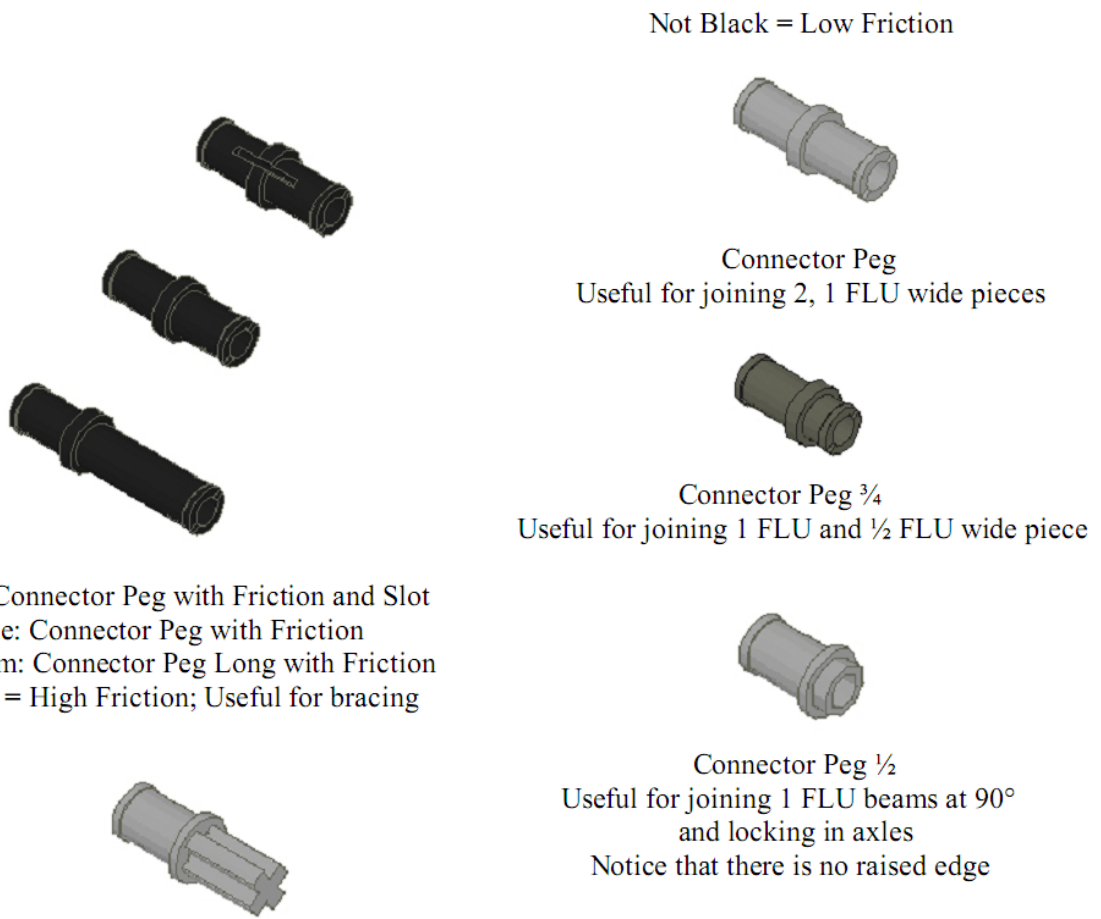

Top: Connector Peg with Friction and Slot
Middle: Connector Peg with Friction
Bottom: Connector Peg Long with Friction
Black = High Friction; Useful for bracing

Not Black = Low Friction

Connector Peg
Useful for joining 2, 1 FLU wide pieces

Connector Peg ¾
Useful for joining 1 FLU and ½ FLU wide piece

Connector Peg ½
Useful for joining 1 FLU beams at 90°
and locking in axles
Notice that there is no raised edge

There are a variety of connector pieces in a variety of colors that have short axle lengths. These can be used for connecting lever arms. . .use your imagination.

### III-C-4 CHANGING ROTATIONAL MOTION: [5]

A train moves by using a piston to drive the first in a chain of wheels. The holes in the gears, besides the central axel catch, can similarly be used to attach an axel that moves almost linearly.

This "piston effect" leads to some interesting ways of harnessing rotational motion:

- Transferring rotational motion over long distances
- A driven, oscillating gear train

The gear rack, in the figure III-C-4-a can be used to change rotational motion into linear motion.

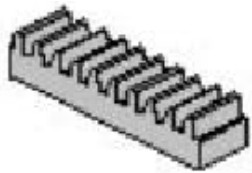

Figure III-C-4-a: gear rack that transforms rotational motion to linear motion.

(Figures credits to [5])

### I-D THE SENSORS:

**Light sensor:**

A sensor that checks the luminosity of the object that it points to, the luminosity varies between in a range of 100 (0→100)

**Collision Sensor:**

The sensor checks whether there is a collision, or not.

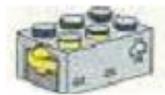 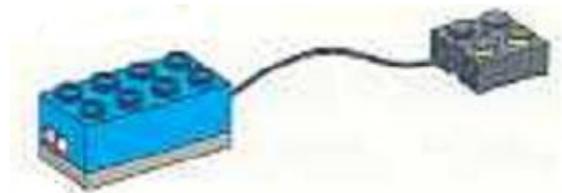

Figure III-D-1: Touch Sensor RIS 2.0.     Figure III-D-2: Light Sensor RIS 2.0.

(Figures credits to [1])

### I-E THE ACTUATORS:

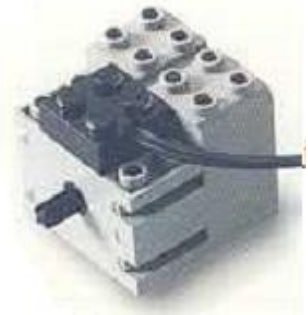

Figure III-E-1: Motor from the KIT RIS 2.0.

(Figures credits to [1])

The motors, or the actuators, are one of the main components of the Kit, since thanks to them we can build moving things. They have 3 kinds of input: rotate forward, rotate backward, or stop.

### SECTION II: PROGRAMMING THE RCX

#### II-A USING THE LEGO® SCRIPT CODE (LSC) :[6]

The RIS2.0 graphical environment is considerably more powerful than that provided with RIS1.5, but it still cannot fully exploit the commands supported by the RCX firmware. In order to open up the full programming power of the RCX firmware, Lego released their scripting language for the RCX, the P-Brick script code language, as part of SDK2.

All LSC script language programs is all have a similar structure to this program. The program is introduced with the statement:

```
program programname {
  // the program goes in here, between the curly brackets
  //(which are also known as braces)
}
```

Inside the program brackets can be one or more tasks that the robot can run at any one time. There must always be a special task defined within the program, known as main { … }. This is the task that is first executed when the program is run.

```
program ProgramName {
main{ //the main task is the one that will first be executed when the
     //  program is run on the brick
    //run the program on the brick by pressing the green Run button
}}
```

In our first example program, we have some definitions, or declarations, that appear before the main task. These definitions will apply to all the tasks in the program. The definitions we have used give names to the output ports so that we can easily refer to them as such in our program. To name the outputs we use the format: `output outputName on N` . Where N is 1, 2 or 3, which correspond to the output ports A, B and C respectively on the brick. Each different instruction should appear on a different line in the program for the sake of clarity. Unlike some other programming languages which use special characters, such as semicolons, to identify the end of a program statement, there are no special characters that mark out the end of line. Within the main task are the instructions that will actually control the robot.

The first instruction in our main task is forward motorA. This sets the direction of the motor on output port A to the forwards direction. (You should make sure the motor is wired correctly so that the forward direction in the program does indeed make the actual motor turn in the direction you want to refer to as forward). The second line – `fd motorC` - sets the direction of the motor on output port C also in the forward direction. **fd** is simply a predefined, shorthand way of writing forward.

The next line, **on** `[motorA motorC]` actually turns on the motors. The square brackets `[ ]` define a list of things that the on command applies to, in this case both of the motors. The motors will stay on in this direction until another motor command is applied to them.

Each instruction in the program executes very quickly. Once a command instruction has been executed, control passes immediately to the next command (although we will see later how the flow of control can be modified using special control structures). If you want the program to take its time going between different instructions, you can use the wait value command. So for example, the wait 100 command tells the program to wait for 100 ticks before going on to the next step. **Each tick lasts for 0.01s (that is one hundredth of a second)**, so wait 100 means wait for 1 second.The **b**ackward `[motorC motor A]` command sets the direction of the two motors to backwards The order in which the outputs appear in the square bracketed list does not matter.The on [motorA motorC] turns on the motors (now in the backwards direction) for 150 ticks (1.5seconds) and then turns them off. This is equivalent to writing:

```
on [motorA motorC]
wait 150
off [motorA motorC]
```

In this case, the **off** `[motorA motorC]` command explicitly turns off the motors. You can change the power level of the motors with the command power **motorA N**, where N is an integer in the range 1 to 8, and where 8 is the maximum power level. The power level should be set within a task itself, for example just before or after the motor direction is set. For example, to set the power level of both motors to 4, use the command `power [motorA motorC]` **4**.

For more information visit: http://robofesta.open.ac.uk/lsc

### II-A USING JAVA WITH LeJOS:

LeJOS is a firmware replacement for Lego Mindstorms programmable bricks. It currently supports the LEGO RCX brick and leJOS NXJ supports the NXT brick. It includes a Java virtual machine, so allows Lego Mindstorms robots to be programmed in the Java programming language. It is often used for teaching Java to first-year computer science students [7]. LeJOS was originally forked out of the TinyVM project. It contains a VM for Java bytecodes and additional software to load and run Java programs. Its features are more flexible than the original OS, since it enables Object oriented language(java), Preemptive threads, Arrays, including multi-dimensional ones, Recursion, Synchronization, Exceptions, Math class, and a Well-documented Robotics API [8].

Sample Code[9]:

```java
///////////////////////////////////
/**
 * Represents a simple sample application.
 *
 * @author The leJOS Tutorial
 * @version 1.0
 */
public class SimpleSample {

    /////////////////////////////////////////
    // public methods
    /////////////////////////////////////////

    /////////////////////////////////////////
    /**
     * main method
     * @throws InterruptedException
     */
    public static void main(String[] args)
        throws InterruptedException {

        // message
        TextLCD.print("DRIVE");

        // drive forward
        Motor.A.forward();

        // just run until RUN button is pressed again
        Button.RUN.waitForPressAndRelease();

    } // main()

} // class SimpleSample
```

For object oriented programmers, LeJOS will be the excellent tool to program their robots, especially when there are complex robot designs that do amazing things.

## SECTION III: DEVELOPMENT AND APPLICATIONS

### III-A COMPARISON WITH THE "NXT KIT":

**Lego Mindstorms NXT** is a programmable robotics kit released by Lego in late July 2006. It replaces the first-generation Lego Mindstorms kit, which was called the Robotics Invention System. The base kit ships in two versions: the Retail Version and the Education Base Set These can be used to compete in FIRST Lego League competitions. It comes with the NXT-G programming software, but a variety of unofficial languages exist, such as NXC, NBC, leJOS NXJ, and RobotC. A new version of the set, the new Lego Mindstorms NXT 2.0, was announced in January 2009, featuring an advanced color sensor and other upgraded capabilities. [10]

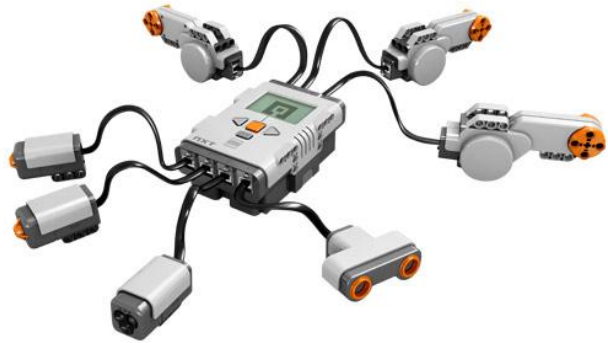

Figure III-A-1: The NXT 2.0 micro-computer attached to different sensors and actuators.

(Figure Credits to [11] )

The Kit holds 619 elements for creating a robots, that includes LEGO TECHNIC building elements, gears, wheels, tracks and tires with a 1 NXT micro-computer, 2 Touch Sensors , 1 Ultrasonic Sensor that makes the robot detect motion , 1 Color Sensor that can detect different colors, light settings and acts as a lamp, 3 Interactive servo motors with built-in rotation sensors, 7 connector cables for linking motors and sensors to the NXT , with a user Guide and a CD that includes an easy-to-use software with an icon-based programming language called NXT-G, and 16 building and programming challenges . The Kit costs 260$ [11]

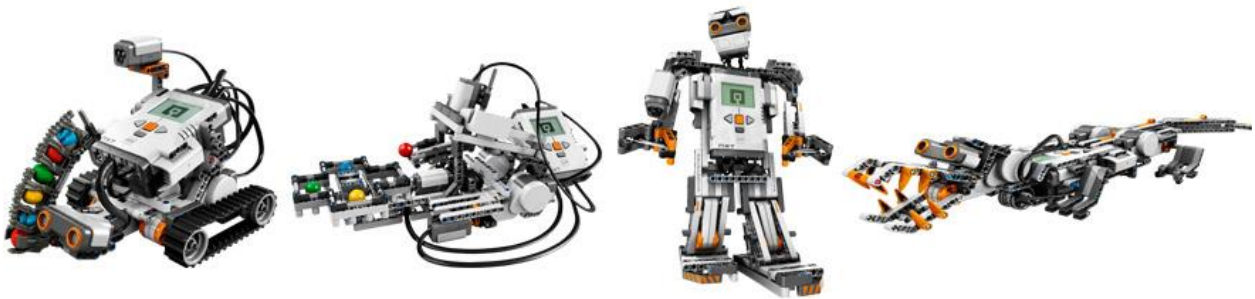

Figure III-A-2: The Different Robot models based on the NXT 2.0 Tool Kit (Figure credits to [11])

### III-B Comparing RCX to "the HandyBoard":

The **Handy Board** is a popular handheld robotics controller. The **Handy Board** was developed at MIT by Fred G. Martin, and was closely based on a previous controller designed by Martin and Randy Sargent for the MIT LEGO Robot Contest. The Handy Board design is licensed free of charge. Thus, several manufacturers make Handy Boards. The Handy Board is used by hundreds of schools worldwide and by many hobbyists for their robot projects. [12]

The Handy Board is based on the 52-pin Motorola MC68HC11 processor, and includes 32K of battery-backed static RAM, four outputs for DC motors, a connector system that allows active sensors to be individually plugged into the board, an LCD screen, and an integrated, rechargeable battery pack. This design is ideal for experimental robotics project, but the Handy Board can serve any number of embedded control applications. [13]

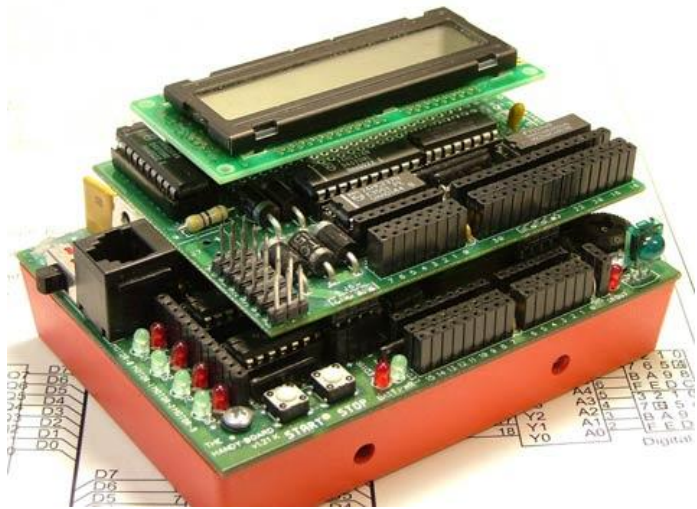

Figure III-B -1: The Handy Board (Figure Credits to [12])

The Handy Board is able to hold [12]:

- 32KB battery-backed SRAM.

- 2x16 LCD character display

- Support for 4 1A motors.

- 6 Servo motor controllers, 7 Digital and 9 Analog inputs, 8 Digital and 16 Analog outputs.

- Infrared I/O capabilities, Serial interface capabilities, and Sound output.

Unlike the RCX, the HandyBoard can hold multiple inputs and outputs. Has more computing power, and could be assembled to a LEGO building easily. Thus, it is a wise choice for strong robotic design. Moreover, it is open source hardware.

### III-C APPLICATION:

#### III-C-1- THE LITTLEBOT:

LittleBot is a robot that I've created, based on the Roverbot model, and I've adjusted some building blocks and features.

The LittleBot Life is only shorted on following the Black lines; whenever it sees a black line it follow the path till the end. The algorithm consists of going forward whenever there is a black line, but when a not black spot is met, then comes the servo part of the algorithm, where the complexity remains. Here is the LSC script the algorithm and shots of the LittleBot and its elements.

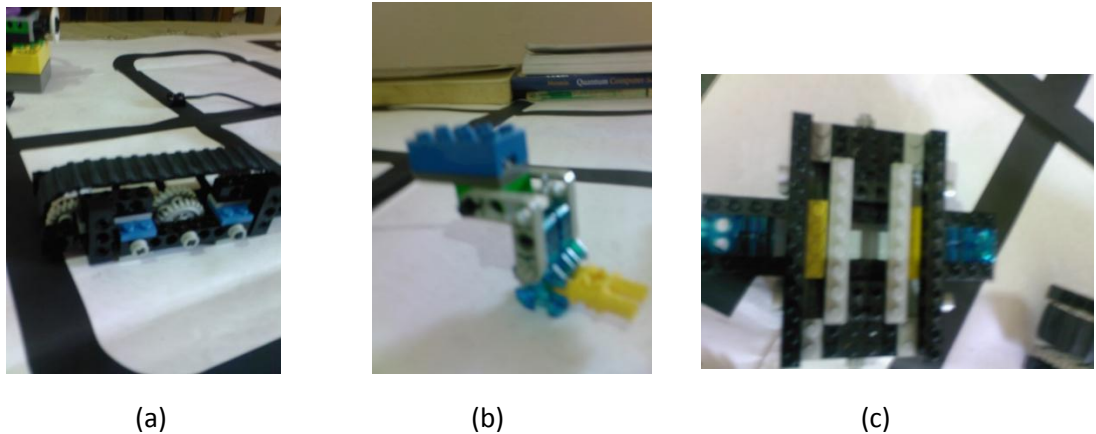

(a)  (b)  (c)

Figure III-C-1: (a) the caterpillars with their gears (b) the light sensor base (c) the robot base structure

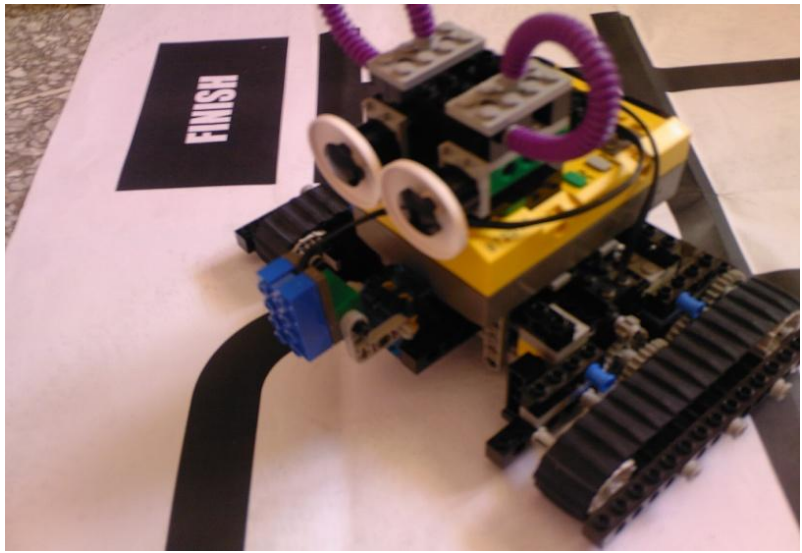

Figure III-C-2: The Final assembly of the robot

(Figures Credit to Ali Elouafiq)

```
program test {
#include <RCX2.h>
#include <RCX2MLT.h>
#include <RCX2Sounds.h>
#include <RCX2Def.h>
var bool = 0
var trig = 0
var mov = 0
var cnt = 0
sensor light2 on 2
light2 is light as percent
macro INIT {
      bool = 0
      trig = 0
      mov = 0
      cnt = 0
}
main {
      ext InterfaceType "kRoverBot"
      rcx_ClearTimers
      bbs_GlobalReset([A B C])
      try {
            INIT
      } retry on fail
      try {
          while timer1 < 3000 {
              if light2 < 52{
                  bb_Forward(A, C, 10)
                  cnt = 0
                  mov = 0
              }
              else
              {
                  if bool = 0{
                      if cnt < 4{
                          bb_SpinLeft(A, C, 10)
                          trig = 0
                          cnt += 1
                      }
                      else
                      {
                          if mov > 6{
```

```
                                bb_SpinLeft(A, C, 10)
                                    mov = 0
                                    cnt = -1
                                bool = 0
                                }
                    else
                        {
                            bb_SpinRight(A, C, 10)
                            mov += 1
                            bool = 10
                        }
                    }
                }
                else
                {
                    if mov > 6{
                        bb_SpinLeft(A, C, 10)
                        mov = 0
                        cnt = -1
                        bool = 0
                    }
                    else
                    {
                        bb_SpinRight(A, C, 10)
                        mov += 1
                        bool = 10
                    }
                }
            }
        }
    } retry on fail
    }
}
```

Code Written in LSC that describes the algorithm that rules the little Bot
(Code credits to Ali Elouafiq)

### III-C-2- THE ARMBOT:

ArmBot is a robotic arm that I've created by facing the robotic pro challenges of the Mindstorms guide, and after doing a research on robotic arms and their design. The ArmBot has 2 DOFs (Degrees of Freedom, not including the gripper), one gripper with one degree of freedom that enables the Arm to grab objects. The arm consists of two actuators, and one light sensor that permit to the arm to detect stuff to grab. The Arm keeps rotating looking for something, when it finds something it takes it and put it in the start point of the arm, and continue its journey grabbing things. The Arm workspace is constituted by the Circle, which center is the center of the arm, and the Radius is 20cms. In Mathematical notation, the circle that has the equation: $x^2 + y^2 = 400 \; cm^2$. So the arm could only grab objects in the coordinates that have the following properties: $(x, y) = (20 \cos\theta, 20 \sin\theta)$

The kinematic properties of the arm are as follow:

$$T = 22.5 \; s \; ; V = 5.6 \cdot 10^{-2} \frac{m}{s}; \omega = 0.279 \frac{rad}{s} \; ; f = 0.044 \; Hz$$

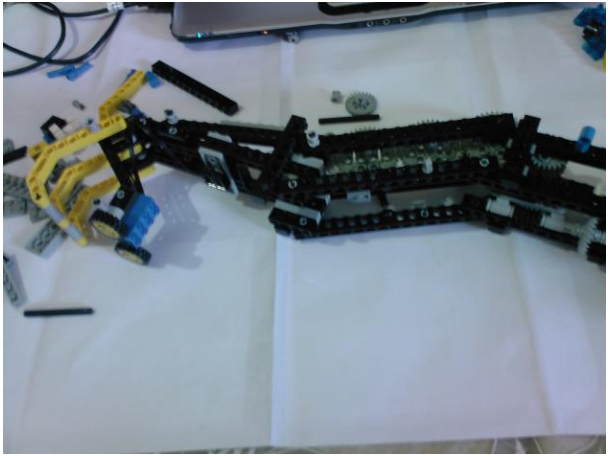
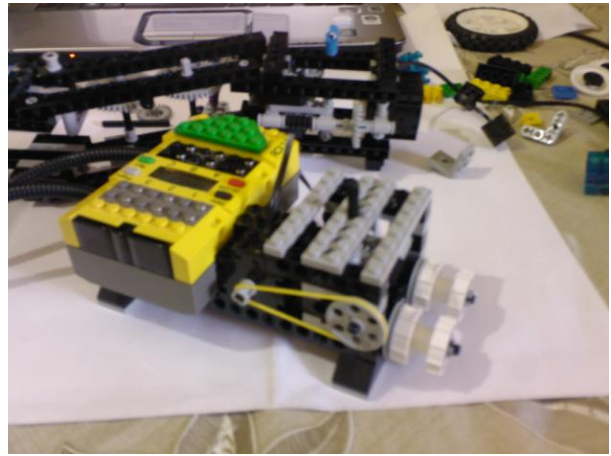

Figure III-C-2-a: The Base structure at the right, the arm at the left (Figures credits to Ali Elouafiq)

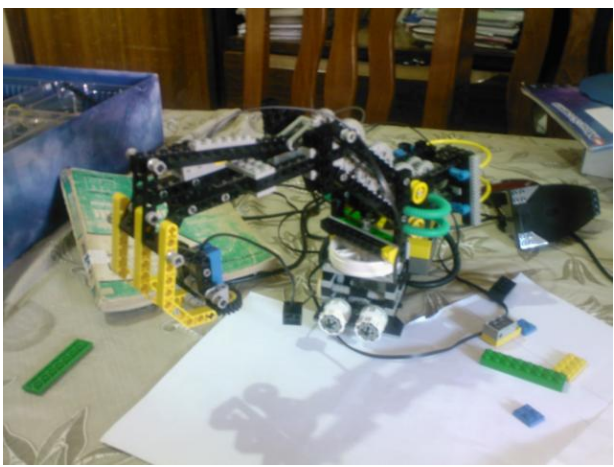
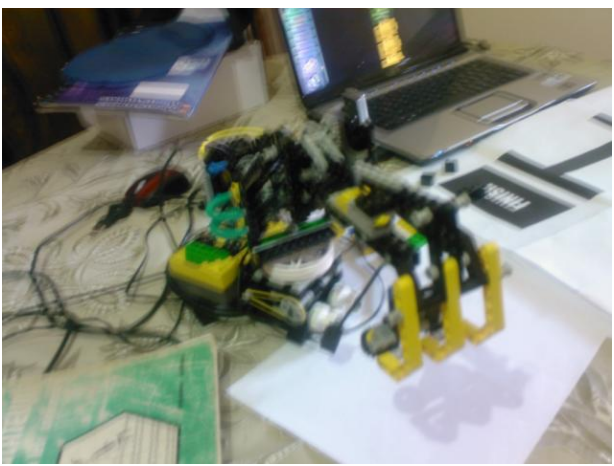

Figure III-C-2-b: The Base Finalized Model of the ArmBot. (Figures credits to Ali Elouafiq)

The ArmBot algorithm's, implanted in LSC: screen shot of the code:

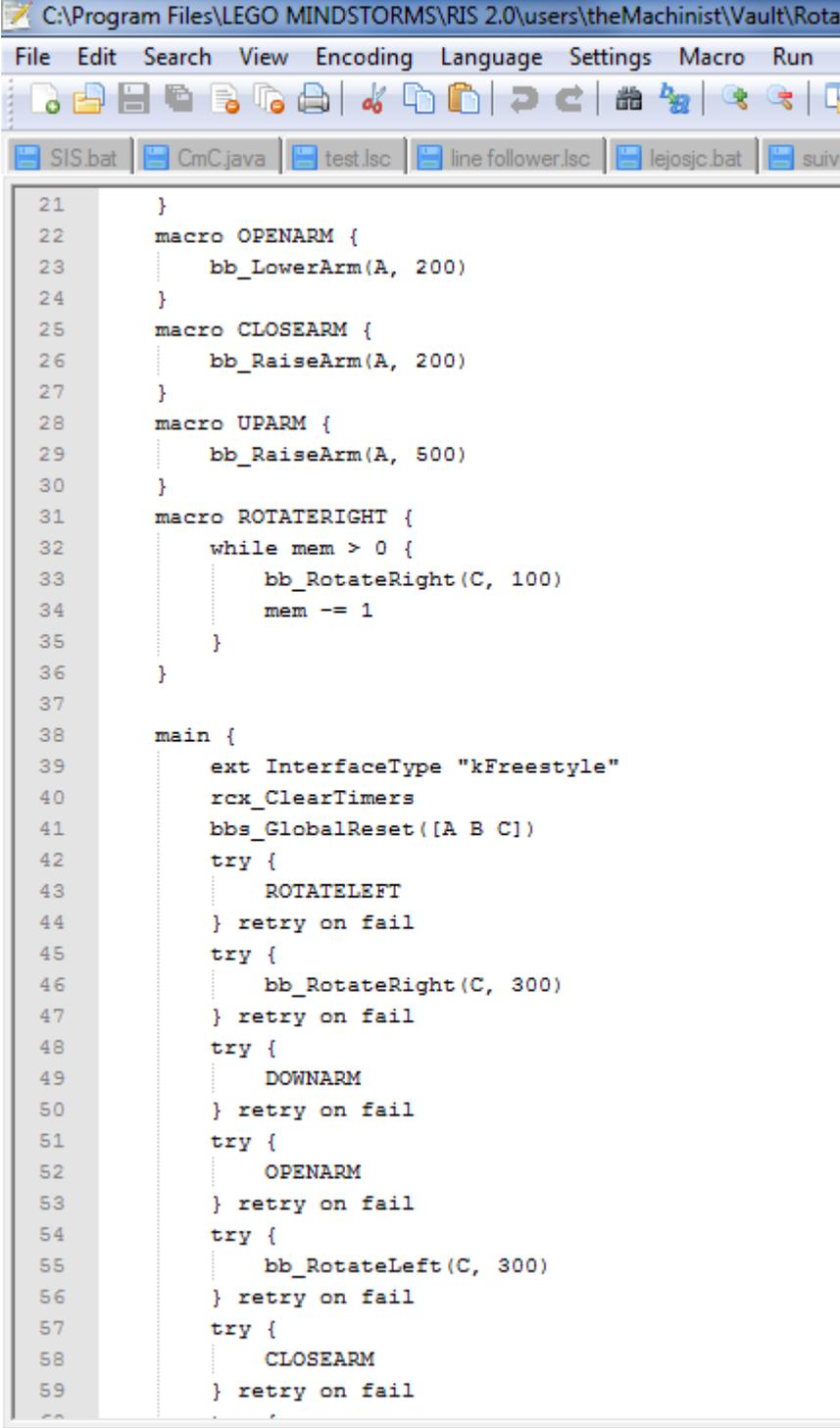

Figure III-C-2-c: the Source code of the ArmBot.

(Code Credits and Figures Credit to Ali Elouafiq)

## Conclusion:

After using and evaluating the LEGO® MINDSTORMS ™ RIS2.0™ tool kit, I've found that it's a versatile robot building tool. To start with its building blocks that enable to build various construction and all kinds of forms for strong robots. Moreover, the RCX2.0 micro computer could be programmed using the Lego Script Code, or using JAVA through LeJOS (a Tiny Virtual Machine) for object oriented programmers. However, there are new versions of the MINDSTORMS ™(notably NXT and NXT 2.0) that have multiple sensors and more building blocks and advanced SDKs, without forgetting the large community of developers. Besides that, more powerful microcomputers are available, such as the MIT's Handy board that can hold multiple sensors and multiple actuators, for more flexibility in robotic design. Finally, we can deduce that the LEGO® MINDSTORM™ Toolkit is a very powerful learning and prototyping tool kit that can be combined easily with various technologies, thereby, leading to a rich learning or prototype testing experience.